\documentclass{article}
\usepackage{ijcai20}

\usepackage{times}

\usepackage{soul}
\usepackage{url}
\usepackage[hidelinks]{hyperref}
\usepackage[utf8]{inputenc}
\usepackage[small]{caption}
\usepackage{graphicx}
\usepackage{amsmath}
\usepackage{booktabs}
\usepackage{amsfonts,amsmath,balance}
\newcommand*{\Scale}[2][4]{\scalebox{#1}{$#2$}}
\title{Canonicalizing Open Knowledge Bases with Multi-Layered \\ Meta-Graph Neural Network}

\author{Tianwen Jiang$^{1,2}$\and
Tong Zhao$^{2}$\and
Bing Qin$^{1}$\and
Ting Liu$^{1}$\and
Nitesh V. Chawla$^{2}$\and 
Meng Jiang$^{2}$
\affiliations
$^1$Research Center for Social Computing and Information Retrieval, Harbin Institute of Technology, China\\
$^2$Department of Computer Science and Engineering, University of Notre Dame, USA\\
\emails
twjiang@ir.hit.edu.cn, tzhao2@nd.edu, \{bqin, tliu\}@ir.hit.edu.cn, \{nchawla, mjiang2\}@nd.edu
}

\begin{document}

\maketitle

\begin{abstract}
Noun phrases and relational phrases in Open Knowledge Bases are often not canonical, leading to redundant and ambiguous facts. In this work, we integrate structural information (from which tuple, which sentence) and semantic information (semantic similarity) to do the canonicalization. We represent the two types of information as a multi-layered graph: the structural information forms the links across the sentence, relational phrase, and noun phrase layers; the semantic information forms weighted intra-layer links for each layer. We propose a graph neural network model to aggregate the representations of noun phrases and relational phrases through the multi-layered meta-graph structure. Experiments show that our model outperforms existing approaches on a public datasets in general domain. 
\end{abstract}

\section{Introduction}
Open Knowledge Bases (Open KBs) do not require specification of ontology or relational schema, and thus can easily adapt to new domains or new data. They were named for being constructed by open information extraction (Open IE) systems such as ReVerb~\cite{fader2011identifying}, OLLIE~\cite{Saha2017BootstrappingFN}, and many others~\cite{angeli2015leveraging,stanovsky2018supervised}. However, Gal\'{a}rraga \textit{et al.}~\shortcite{galarraga2014canonicalizing} argue that the noun phrases (NPs) and relational phrases (RPs) in Open KBs are often not canonical as they may have various forms and can hardly be linked to standard KBs (e.g., Freebase, Wikipedia). For example, ``CIFS (Common Internet File System)'' is a general-purpose information-sharing protocol formerly known as ``SMB (Server essage Block)'', however, Open KBs may often treat them as totally different NPs directly. The problem of canonicalization is to group NPs (those referring to the same entity) and RPs (those having the same semantic meaning) in the given Open KBs.

\begin{figure}[t]
    \centering
    \includegraphics[width=1.0\linewidth]{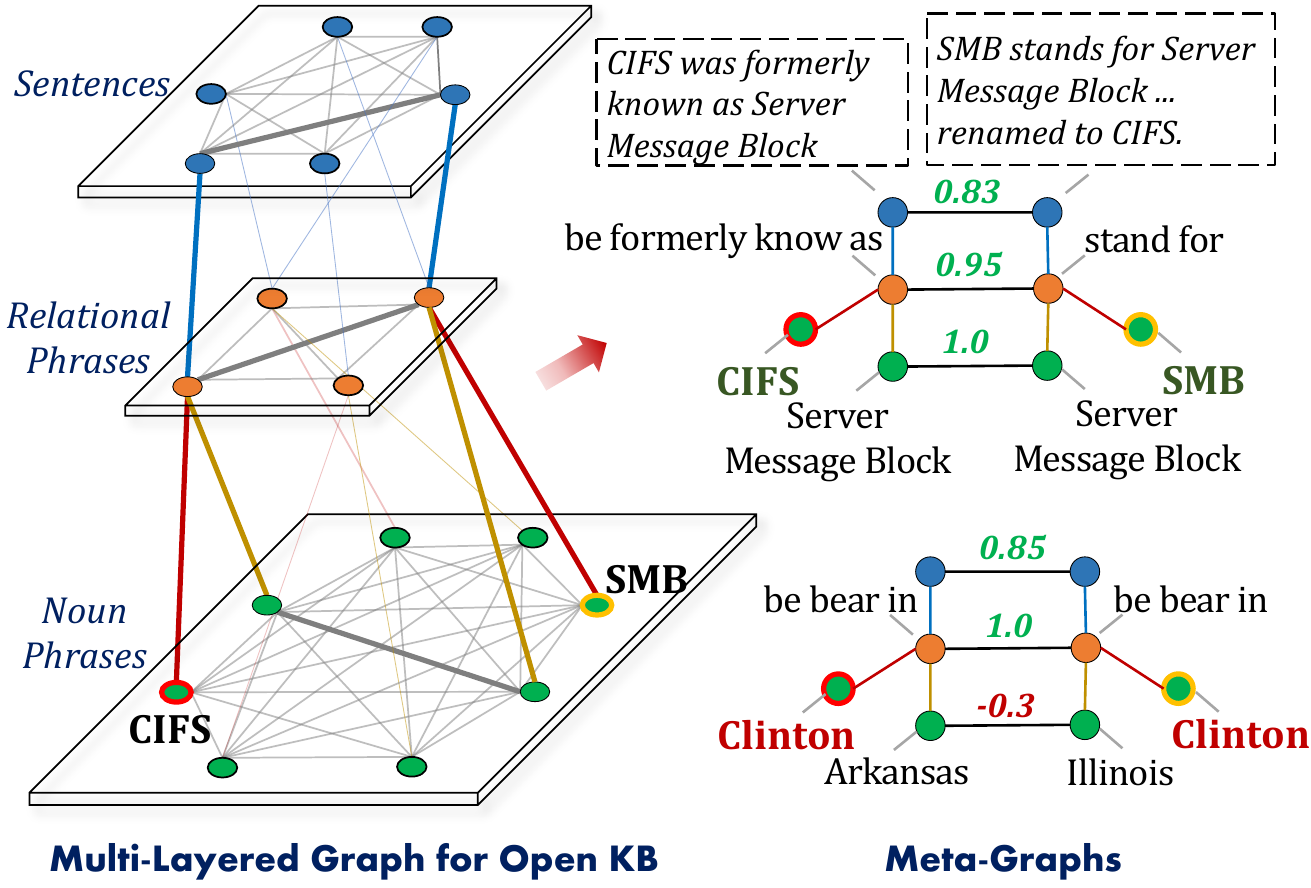}
    \vspace{-0.2in}
    \caption{An Open KB can be represented as a three-layered graph. We propose to address the NP and RP canonicalization problem by learning node representations via graph neural network based on meta-graph structures.}
    \label{fig:idea}
    \vspace{-0.1in}
\end{figure}

\begin{figure*}[t]
    \centering
    \includegraphics[width=1.0\linewidth]{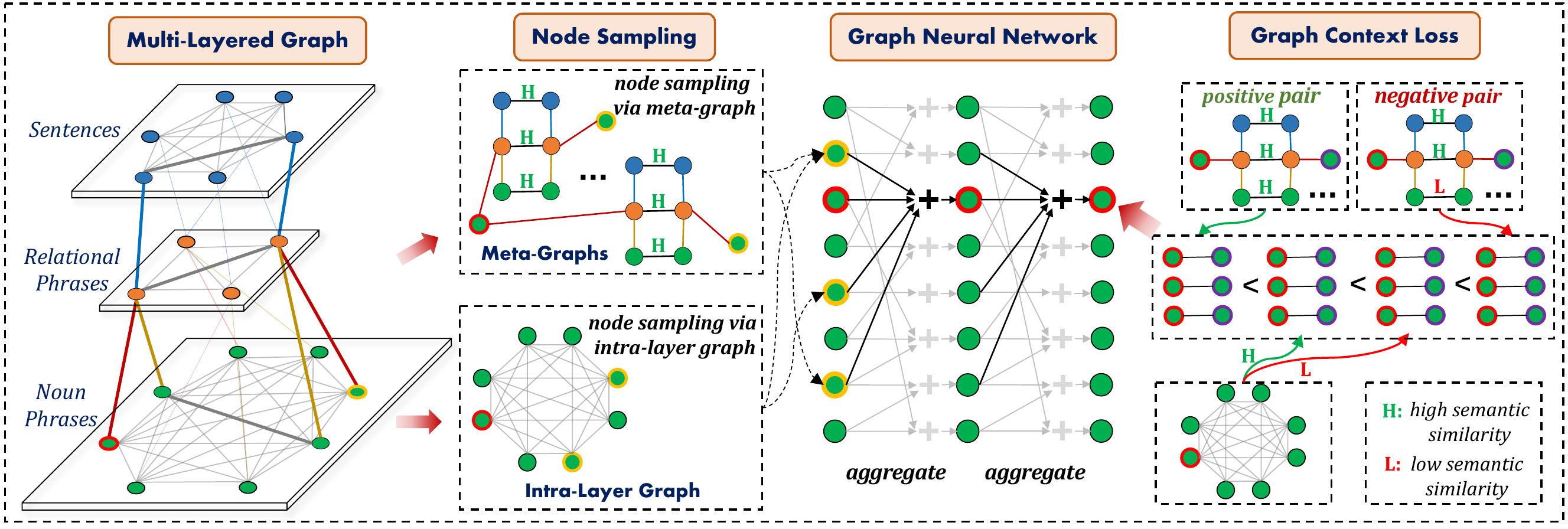}
    \vspace{-0.2in}
    \caption{The framework of multi-layered meta-graph neural network.}
    \label{fig:framework}
    \vspace{-0.1in}
\end{figure*}

Existing methods use text embedding algorithms~\cite{pennington2014glove,mikolov2013distributed} assuming that
the NPs (RPs) with similar contexts or similar surface forms, e.g., ``city\_of\_chicago'' and ``chicago\_area'', ``be\_bear\_in'' and ``be\_bear\_at'', can be grouped together, The semantic embeddings perform better than feature engineering methods~\cite{vashishth2018cesi}. However, as the examples in Fig.~\ref{fig:idea}, (a) the NPs or RPs may look different but actually they refer to the same thing in the two tuples: (``CIFS'', be\_formally\_known\_as, Server\_Message\_Block) and (``SMB'', stand\_for, Server\_Message\_Block); (b) the same surface may refer to different things: (``Clinton'', be\_bear\_in, Arkansas) and (``Clinton'', be\_bear\_in, Illinois). We argue that a better sense of the NP/RP ambiguity needs integration of (1) sentence-to-tuple-to-phrase structural information and (2) semantic information of contexts.

We conduct the Open KB as a three-layered graph (see Fig.~\ref{fig:idea}). The layers from bottom to top are of NP nodes, RP nodes, and sentence nodes. The structural information forms inter-layer links connecting NPs to the RP (with attributes ``subject'' and ``object'' as a tuple) and RPs to the sentences. The semantic information forms intra-layer links for each layer, weighted by the semantic similarity between the nodes. The semantic embedding can be obtained from GloVe, language models, or some open IE methods. In such three-layered knowledge graph, distinguishing (dis-)similar nodes for a given node is essential for canonicalization problem. Therefore we propose a novel structure, called \textit{multi-layered meta-graph}, to collectively sample (dis-)similar nodes for Open KB canonicalization.

A multi-layered meta-graph is an induced sub-graph of inter-layer and intra-layer links. Various meta-graph can be defined, while a basic meta-graph is composed of all the nodes and links from \textit{two} Open KB relation tuples. The (dis-)similar nodes can be selected and sampled via multi-layered meta-graph structures. For example (see Fig.~\ref{fig:idea}), the meta-graph connecting ``CIFS'' and ``SMB'' has high positive weights on intra-layer links between sentences, RPs, and NPs -- it indicates similarity; the meta graph connecting the two ``Clinton'' nodes has high positive weights on sentence and RP links but negative weights on the NP-NP link -- it indicates dissimilarity. Note that most of the meta graphs are not strong indicators.

We propose a new graph neural network model, multi-layered meta-graph neural network (MGNN), to learn \textit{canonical embedding} based on semantic information (induced into semantic embedding) and structural information (induced into meta-graph). Two phrases (i.e., NPs or RPs) of the similar canonical embeddings should be grouped together. Given one NP or RP, MGNN first concatenates the embeddings of its similar NPs or RPs sampled based on semantic embedding similarity and meta-graph structures. MGNN then aggregate these sampled nodes with a non-linear weighted transform of the concatenated embedding to update the canonical embedding of the target NP or RP node. The learning process is unsupervised, and we introduce a hybrid loss function to guide MGNN for effective canonical embedding learning.



We conduct experiments on a public Open KB (from ReVerb)~\cite{fader2011identifying}. Results show that our model outperforms existing approaches on OpenKB canonicalization.

\section{Problem Definition}
Open IE systems extract (subject, relation, object)-tuples from sentences to build Open KBs. Here we put the sentence into the tuple, as the task of canonicalization needs the full context. We define a new tuple representation:

\noindent\textbf{Definition 1 (Open KB Relation Tuple)} A (sentence, relation, subject, object)-tuple is used to describe the relation tuple extraction, denoted by $(s, r, e^{subj}, e^{obj})$, where $s\in\mathcal{S}, r\in\mathcal{R}, e^{subj}, e^{obj}\in\mathcal{E}$. Here $\mathcal{S}, \mathcal{R}, \mathcal{E}$ denote the set of sentences, relational phrases, and noun phrases in Open KB. The subscript ``subj'' is for the subject role and ``obj'' is for object. The traditional tuple is $(e^{subj}, r, e^{obj})$.

\noindent\textbf{Definition 2 (Canonical Embedding)} For a phrase, either an NP or RP, $x \in \mathcal{E} \cup \mathcal{R}$, the canonical embedding is denoted by $\textbf{z}^x \in \mathbb{R}^{d}$, where $d$ is the number of latent features. Ideally, two phrases of similar canonical embeddings should be grouped together and those of dissimilar embeddings should be separated. These embeddings will be used in a standard way \cite{galarraga2014canonicalizing,vashishth2018cesi}, say, fed into clustering algorithms for Open KB canonicalization.

\noindent\textbf{Problem (Open KB Canonicalization)} Given (a) the structural information denoted by a set of open KB relation tuples $\mathcal{D} = \{(s, r, e^{subj}, e^{obj})\}$ and (b) the semantic information denoted by $\mathbf{v}^x \in \mathbb{R}^{d_0}$ ($\forall x \in \mathcal{E} \cup \mathcal{R} \cup \mathcal{S}$), find the canonical embedding $\textbf{z}^x$ for $x \in \mathcal{E} \cup \mathcal{R}$, and then apply clustering algorithm to group NPs (those refer to the same entity) and RPs (those have the same semantic) for the Open KB. Here $\mathbf{v}^x$ is the phrase or sentence's semantic feature vector given by GloVe, language modes, or the open IE systems; $d_0$ is the number of dimensions, usually $d_0 \geq d$.

\section{The Proposed Model}
The model design of our proposed multi-layered meta-graph neural network (MGNN) is shown in Fig.~\ref{fig:framework}. Open KBs are represented as multi-layered graphs. The structural information forms inter-layer links and the semantic information forms intra-layer links. MGNN learns the node's canonical embeddings through the graph neural architecture, guided by semantic information (induced into semantic embedding) and structural information (induced into meta-graph). After that, we apply Hierarchical Agglomerative Clustering (HAC)~\cite{tan2006introduction} on the learned canonical embeddings of NPs and RPs to obtain the canonicalization for Open KBs.

\subsection{Multi-Layered Graph for Open KBs}
We use multi-layered graph instead of the traditional relational graphs (flatten network of entities) to represent Open KBs: $\mathcal{G} = \{L_1, L_2, L_3, E^{subj}_{1,2}, E^{obj}_{1,2}, E_{2,3}\}$. $L_i$ denotes the $i$-th layer (from bottom to top). All the three layers have fully linked intra-layer links, inducing the semantic information from pre-learned semantic embeddings (e.g., word2vec, GloVe). $E_{i,j}$ denotes the inter-layer links. We use $v^x$ represents the corresponding node of relation tuple unit $x\ (x \in \{s, r, e^{subj}, e^{obj}\})$, and $\mathbf{v}^x$ is to denote the semantic embedding of $v^x$. $V_{\mathcal{X}}$ is to denote the nodes sets for $\mathcal{X}$, where $\mathcal{X}$ is a set of NPs, RPs or sentences from $\mathcal{S}, \mathcal{R}$ or $\mathcal{E}$. 

\textbf{{The first layer}} $L_1 = \{V_\mathcal{E}, E_{\mathcal{E}, \mathcal{E}}\}$ for NP nodes. An intra-layer link $(v^{e_i}, v^{e_j})$ exists between any two NP nodes $v^{e_i}$ and $v^{e_j}$. And the weight of the intra-layer link is defined as $\Phi(v^{e_i}, v^{e_j}) = \mathbf{v}^{e_i}\cdot\mathbf{v}^{e_j}$.

\textbf{{The second layer}} $L_2 = \{V_\mathcal{R}, E_{\mathcal{R}, \mathcal{R}}\}$ for RP nodes. An intra-layer link $(v^{r_i}, v^{r_j})$ exists between any two RP nodes $v^{r_i}$ and $v^{r_j}$, and the link weight is $\Phi(v^{r_i}, v^{r_j}) = \mathbf{v}^{r_i}\cdot\mathbf{v}^{r_j}$.

\textbf{{The third layer}} $L_3 = \{V_\mathcal{S}, E_{\mathcal{S}, \mathcal{S}}\}$ for sentence nodes. An intra-layer link $(v^{s_i}, v^{s_j})$ exists between any two sentence nodes $v^{s_i}$, $v^{s_j}$. The link weight is $\Phi(v^{s_i}, v^{s_j}) = \mathbf{v}^{s_i}\cdot\mathbf{v}^{s_j}$.

Between layers, we have sets of \textbf{\textit{inter-layer links}}, where $E^{subj}_{1,2}$ = $\{(v^{e^{subj}},v^r)\ |\ (s, r, e^{subj}, e^{obj}) \in \mathcal{D}\}$, $E^{obj}_{1,2}$ = $\{(v^{e^{obj}},v^r)\ |\ (s, r, e^{subj}, e^{obj})$ $\in \mathcal{D}\}$, $E_{2,3}$ = $\{(v^r,v^s)\ |\ (s,$ $r, e^{subj}, e^{obj}) \in \mathcal{D}\}$.

Note that two NPs of the same surface form (e.g., ``Clinton'') are considered as two different nodes. So, one NP node connects to only one RP node in $\mathcal{G}$. For semantic embeddings, we use bag-of-words followed by SVD for sentences, GloVe for NPs (average vector for phrase), and  BERT~\cite{devlin2018bert} for RPs (average vector for phrase), and then do the normalization for all vectors.

\subsection{Multi-Layered Meta Graph}
We define a new graph structure, called multi-layered meta graph (simply named as meta-graph), which is used to sample (dis-)similar nodes for a given NP/RP node. These (dis-)similar nodes will be utilized to aggregate their features for canonical embedding learning and canonicalization. For convenience, we first define that $E_{\mathcal{G}}=E^{subj}_{1,2}\cup E^{obj}_{1,2}\cup E_{2,3}\cup E_{\mathcal{E}, \mathcal{E}}\cup E_{\mathcal{R}, \mathcal{R}}\cup E_{\mathcal{S}, \mathcal{S}}$, $\Delta(V)=\{(v_i, v_j)\ |\ v_i, v_j\in V, (v_i, v_j)\in E_{\mathcal{G}}\}$, $\mathcal{S}_i^e = \{s\ |\ (s,r_i,e^{subj}_i,e^{obj}_i)\in\mathcal{D}\}$ and $\mathcal{S}_i^r = \{s\ |\ (s,r_i,e^{subj},e^{obj})\in\mathcal{D}\}$.

\begin{figure}[t]
    \centering
    \includegraphics[width=0.8\linewidth]{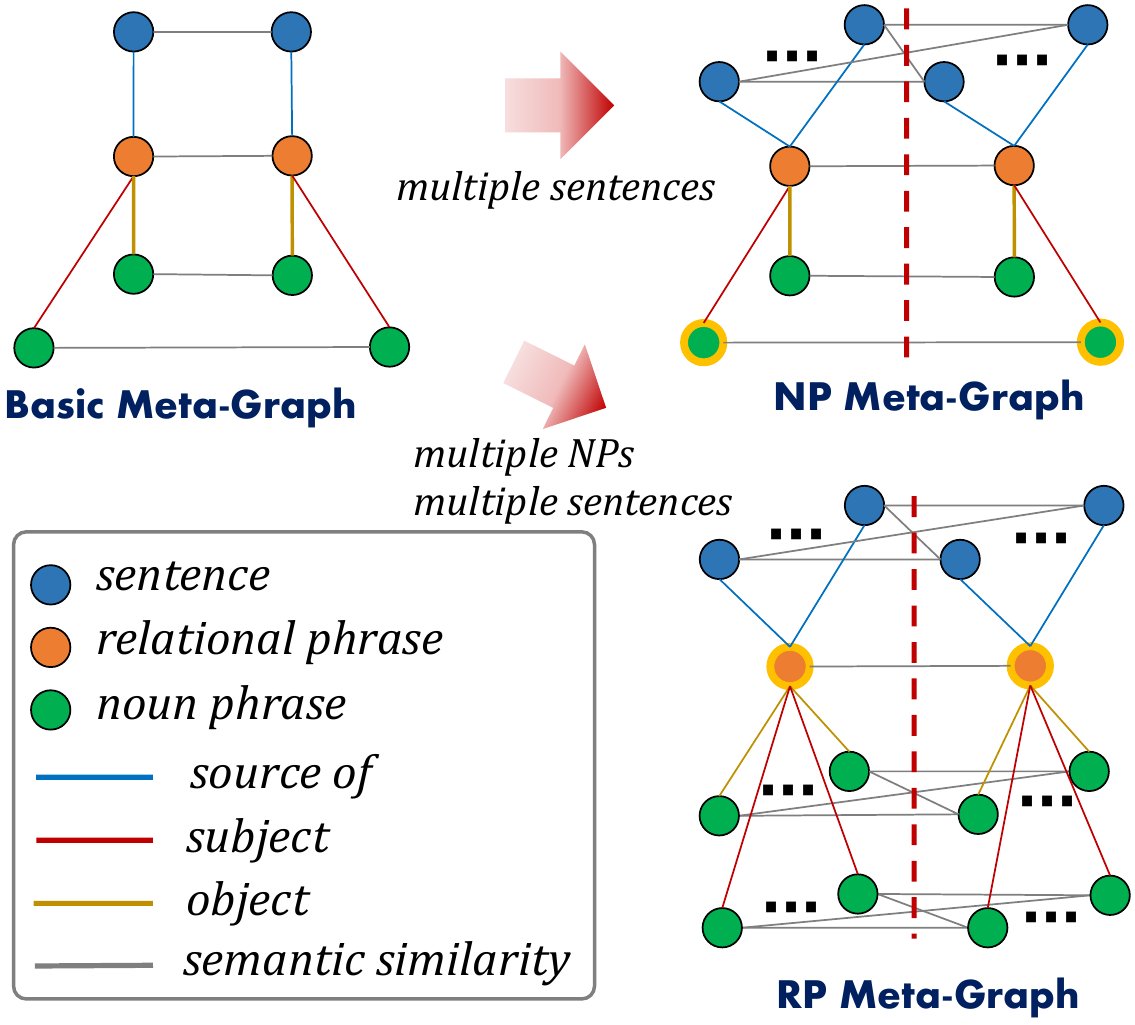}
    \vspace{-0.15in}
    \caption{The meta-graphs connect two NP nodes or RP nodes through the inter-layer and intra-layer links.}
    \label{fig:meta-graph}
    \vspace{-0.1in}
\end{figure}

\noindent \textbf{Basic meta-graph.} As shown in Fig.~\ref{fig:meta-graph}, a basic meta-graph is composed of all the nodes and links from \textit{two} Open KB relation tuples. Therefore a basic meta-graph is defined as $G = (V, \Delta(V))$ where $V=\{v^{s_1}, v^{s_2}, v^{r_1}, v^{r_2}, v^{e^{subj}_1}, v^{e^{subj}_2}, v^{e^{obj}_1}, v^{e^{obj}_2}\}$

\noindent \textbf{NP meta-graph.} For a pair of NP nodes (green), an NP meta-graph is an extension of basic meta-graph for multiple relevant sentences. Formally, given a pair of NP as subjects in certain tuples $e^{subj}_1, e^{subj}_2$, suppose the corresponding RP are ${r_1}, {r_2}$. The NP meta-graph is defined as $G_{{e^{subj}_1}, {e^{subj}_2}} = (V^{NP}, \Delta(V^{NP}))$, where $V^{NP} = V_{\mathcal{S}_1^e}\cup V_{\mathcal{S}_2^e}\cup \{v^{r_1}, v^{r_2}, v^{e^{obj}_1}, v^{e^{obj}_2}\}$. For a pair of object nodes, $V^{NP} = V_{\mathcal{S}_1^e}\cup V_{\mathcal{S}_2^e}\cup \{v^{r_1}, v^{r_2}, v^{e^{subj}_1}, v^{e^{subj}_2}\}$, and so $G_{{e^{obj}_1}, {e^{obj}_2}} = (V^{NP}, \Delta(V^{NP}))$. We simply write the NP meta-graph as $G_{e_1, e_2} = (\hat{V}^{NP}, \Delta(\hat{V}^{NP}))$ where $\hat{V}^{NP} = V_{\mathcal{S}_1^e}\cup V_{\mathcal{S}_2^e}\cup \{v^{r_1}, v^{r_2}, v^{\hat{e}_1}, v^{\hat{e}_2}\}$ for later use.

\noindent \textbf{RP meta-graph.} For a pair of RP nodes (orange), the RP meta-graph includes all the relevant NP nodes and sentence nodes. Formally, suppose the RP pair are $r_1$ and $r_2$. The RP meta-graph is defined as $G_{r_1, r_2}$ = $(V^{RP}, \Delta(V^{RP}))$, where $V^{RP} = V_{\mathcal{S}_1^r}\cup V_{\mathcal{S}_2^r}\cup V_{\mathcal{E}^{subj}_1}\cup V_{\mathcal{E}^{subj}_2}\cup V_{\mathcal{E}^{obj}_1}\cup V_{\mathcal{E}^{obj}_2}$, $\mathcal{E}^{subj}_i = \{e^{subj}\ |\ (s,r_i,e^{subj},e^{obj})\in\mathcal{D}\}$ and $\mathcal{E}^{obj}_i = \{e^{obj}\ |\ (s,r_i,e^{subj},e^{obj})\in\mathcal{D}\}$ ($i=1, 2$).

\subsection{Phrase (Node) Pair Sampling via Meta-Graph}\label{sec:phrasepair}

In this section, we define the \textit{canonical weight} $\Psi(e_1, e_2)$ between a pair of phrases (NP/RP nodes in graph) using their meta-graphs. Higher canonical weight means higher probability to group the phrase (node) pair together. Given a pair of NP nodes $v^{e_1}$ and $v^{e_2}$ with the NP meta-graph $G_{e_1, e_2}$. We define the canonical weight as the mean of the link weight between the sentence node sets ($V_{\mathcal{S}_1}$ and $V_{\mathcal{S}_2}$), RP nodes ($v^{r_1}$ and $v^{r_2}$), and NP nodes ($v^{\hat{e}_1}$ and $v^{\hat{e}_2}$, $v^{e_1}$ and $v^{e_2}$):
\begin{equation}
    \begin{aligned}
    \Psi(v^{e_1}, v^{e_2}) & = \frac{1}{4} ( \hat{\Phi}(V_{\mathcal{S}_1^e}, V_{\mathcal{S}_2^e}) + \Phi(v^{r_1}, v^{r_2}) \\
    & + \Phi(v^{\hat{e}_1}, v^{\hat{e}_2})+\Phi(v^{e_1}, v^{e_2}) )
    \end{aligned}
\end{equation}
where $\Psi(\cdot)$ is the canonical weight function and the link weight between two sentence sets is defined as follows,
\begin{equation}
    \Scale[1.0]{
    \hat{\Phi}(V_{\mathcal{S}_1^e}, V_{\mathcal{S}_2^e}) = \frac{\sum\limits_{s_1 \in \mathcal{S}_1^e} \sum\limits_{s_2 \in \mathcal{S}_2^e} \Phi({v}^{s_1}, {v}^{s_2})}{|\mathcal{S}_1^e| \cdot |\mathcal{S}_2^e|} = \frac{\sum\limits_{s_1 \in \mathcal{S}_1} \sum\limits_{s_2 \in \mathcal{S}_2^e} \mathbf{v}^{s_1}\cdot\mathbf{v}^{s_2}}{|\mathcal{S}_1^e| \cdot |\mathcal{S}_2^e|}.}
\end{equation}

On the other hand, given a pair of RP nodes $v^{r_1}$ and $v^{r_2}$, with the RP meta-graph $G_{r_1, r_2}$. The canonical weight between the RP nodes is
\begin{equation}
    \begin{aligned}
    & \Psi(v^{r_1}, v^{r_2}) = \frac{1}{4} (\hat{\Phi}(V_{\mathcal{S}_1^r}, V_{\mathcal{S}_2^r}) + \Phi(v^{r_1}, v^{r_2}) \\
    & + \hat{\Phi}(V_{\mathcal{E}^{subj}_1}, V_{\mathcal{E}^{subj}_2})+\hat{\Phi}(V_{\mathcal{E}^{obj}_1}, V_{\mathcal{E}^{obj}_2})).
    \end{aligned}
\end{equation}
where $\hat{\Phi}(V_{\mathcal{S}_1^r}, V_{\mathcal{S}_2^r})$ is defined same as $\hat{\Phi}(V_{\mathcal{S}_1^e}, V_{\mathcal{S}_2^e})$ by replacing $S_i^e$ with $S_i^r$ ($i=1, 2$). The link weight between two NP sets is defined as
\begin{equation}
    \Scale[1.0]{
    \hat{\Phi}(V_{\mathcal{E}_1}, V_{\mathcal{E}_2}) = \frac{\sum\limits_{e_1 \in \mathcal{E}_1} \sum\limits_{e_2 \in \mathcal{E}_2} \Phi({v}^{e_1}, {v}^{e_2})}{|\mathcal{E}_1| \cdot |\mathcal{E}_2|} = \frac{\sum\limits_{e_1 \in \mathcal{E}_1} \sum\limits_{e_2 \in \mathcal{E}_2} \mathbf{v}^{e_1}\cdot\mathbf{v}^{e_2}}{|\mathcal{E}_1| \cdot |\mathcal{E}_2|}.}
\end{equation}


Next we discuss about finding \textit{negative} phrase pairs for negative sampling in modeling training process. For example, the two NPs ``Clinton'' actually refer to different persons. In their NP meta-graph, we find that the intra-layer links on sentences, RPs, and subject NPs are positively high but the link between the object NPs (``Arkansas'' and ``Illinois'') is negative. So we define the \textit{negative canonical probability} for a pair of NP nodes $v^{e_1}$ and $v^{e_2}$, where higher one means higher probability of NPs referring to different things:
\begin{equation}
    \Psi^{-}(v^{e_1}, v^{e_2}) = \sigma(\frac{-2 \cdot \Phi(v^{\hat{e}_1}, v^{\hat{e}_2})}{\hat{\Phi}(V_{\mathcal{S}_1^e}, V_{\mathcal{S}_2^e}) + \Phi(v^{r_1}, v^{r_2})}),
\end{equation}
where we use $\Psi^{-}(\cdot)$ as a function to negative canonical probability and $\sigma$ is a sigmoid function. Given a meta-graph of a NP pair (e.g., both as subjects), when we found extremely high similarity in sentence pair and RP pair while extreme low similarity in the other NP pair (both as objects), it indicates high probability of the NP pair referring to different things (i.e., negative NP pair).

Similarly, given the RP meta-graph of $v^{r_1}$ and $v^{r_2}$, the function to negative canonical probability is
\begin{equation}
    \Scale[0.9]{\Psi^{-}(v^{r_1}, v^{r_2}) = \sigma(\frac{-2 \cdot \mathrm{min}(\hat{\Phi}(V_{\mathcal{E}^{subj}_1}, V_{\mathcal{E}^{subj}_2}), \hat{\Phi}(V_{\mathcal{E}^{obj}_1}, V_{\mathcal{E}^{obj}_2}))}{\hat{\Phi}(V_{\mathcal{S}_1^r}, V_{\mathcal{S}_2^r}) + \mathrm{max}(\hat{\Phi}(V_{\mathcal{E}^{subj}_1}, V_{\mathcal{E}^{subj}_2}), \hat{\Phi}(V_{\mathcal{E}^{obj}_1}, V_{\mathcal{E}^{obj}_2}))}).}
\end{equation}

\noindent\textbf{Complexity analysis.} It takes $O(N^2)$ time to compute canonical weights of all NP pairs and RP pairs. In practice, we first sort the similarity between sentences, RPs and NPs. Then we adopt early-stop strategy to refuse to calculate the canonical weight of the rest pairs. It cost a reasonable time in fact even has a $O(N^2)$ time complexity.

\subsection{Canonical Embedding Aggregation in Multi-Layered GNN}
We extend GraphSAGE~\cite{hamilton2017inductive}, a sampling and aggregation-based GNN model to the multi-layered graph settings. Given a phrase $x\in \mathcal{E}\cup \mathcal{R}$ along with its node $v^x \in V_\mathcal{E} \cup V_\mathcal{R}$, we initialize the canonical embedding with the semantic embedding:
\begin{equation}
\mathbf{h}^0_x \leftarrow \mathbf{v}^x.
\end{equation}
The next step is to weighted sample a set of $v^x$'s ``neighboring nodes'' $\mathcal{N}(v^x) = \mathcal{N}^{\Psi}(v^x) \cup \mathcal{N}^{\Phi}(v^x)$, where $\mathcal{N}(v^x)$ includes the node samples from both the meta-graph based canonical weight distribution (obtained by $\Psi(v^x,\cdot)$) and intra-layer weight distribution (obtained by $\Phi(v^x,\cdot)$).

Suppose $\mathbf{h}^{k-1}_x$ is the canonical embedding of node $v^x$ at the (k-1)-th GNN layer. We define
\begin{equation}
    \mathrm{\textbf{M}}^{k-1}_x=[\mathrm{\textbf{h}}^{k-1}_{y}, \forall y\in\mathcal{N}(v^x)] \in \mathbb{R}^{|\mathcal{N}(v^x)|\times d},
\end{equation}
whose rows are the embeddings of neighbor node samples. Then, we apply mean pooling aggregators to transform it into a $d$-dimensional vector for the aggregated $v^x$'s neighboring information at the $k$-th GNN layer:
\begin{equation}
    \mathrm{\textbf{h}}^k_{\mathcal{N}(v^x)}=\mathbf{\textsc{aggregate}}_k(\mathrm{\textbf{M}}^{k-1}_x).
\end{equation}
Different GNN layers may choose different aggregators. The next step is to concatenate the node's feature vector at the ($k-1$)-th layer and the neighboring feature vector at the $k$-th layer, and to multiply with a weighted matrix $\mathbf{W}^k$ ($\mathbf{W}^k \in \mathbb{R}^{2d_0\times d}$ if $k=1$; otherwise, $\mathbf{W}^k \in \mathbb{R}^{2d\times d}$), do a non-linear transform with $\sigma$ and normalization.
\begin{equation}
    \begin{aligned}
    \mathrm{\textbf{h}}^k_x&=\sigma\left(\mathrm{\textbf{W}}^k\cdot\mathbf{\textsc{concat}}(\mathrm{\textbf{h}}^{k-1}_x, \mathrm{\textbf{h}}^k_{\mathcal{N}(v^x)})\right),\\
   \mathrm{\textbf{h}}^k_x&=\mathrm{\textbf{h}}^k_x/||\mathrm{\textbf{h}}^k_x||_2.
   \end{aligned}
\end{equation}

After $K$ iterations, the final canonical embedding is denoted by $\mathrm{\textbf{z}}^x\equiv\mathrm{\textbf{h}}_x^K$. Finally, we apply the Hierarchical Agglomerative Clustering (HAC) on $\mathbf{z}^x, x \in \mathcal{E}\cup\mathcal{R}$ for open KB canonicalization.

\subsection{Hybrid Loss in Multi-Layered GNN}

We introduce the loss function to supervise the multi-layered GNN for effective canonical embedding learning. To design the loss, we have the following assumptions: (1) As defined before, a pair of phrases, either NPs or RPs, would have {similar} canonical embeddings if they have {high} canonical weight $\Psi(v^{e_1},v^{e_2})$, $\Psi(v^{r_1},v^{r_2})$, because the meta-graph structure supports their grouping. (2) A pair of phrases should have {dissimilar} canonical embeddings if they have {high} negative canonical probability $\Psi^-(v^{e_1},v^{e_2})$, $\Psi^-(v^{r_1},v^{r_2})$, because the meta-graph structure supports the separation. (3) Besides the meta-graph structures, intra-layer link weights on $L_1$ or $L_2$ (semantic similarity) indicate the grouping/separation: {positive, high} link weight indicates grouping and thus similar canonical embeddings, and {negative, low} link weight indicates separation and dissimilar canonical embeddings. (4) Due to the phrase ambiguity, the pure semantic similarity is not as trustworthy as the meta-graph structure, and thus generates high recall but low precision. So, if we find four sets of phrase pairs:
\begin{itemize}
\item $D_{meta,+}$: its phrase pairs $(x,y)$ have high canonical weight $\Psi(v^x, v^y)$;
\item $D_{meta,-}$: its phrase pairs $(x,y)$ have high negative canonical probability $\Psi^{-}(v^x, v^y)$;
\item $D_{intra,+}$: its phrase pairs $(x,y)$ have positive, high semantic similarity $\Phi(v^x, v^y)$;
\item $D_{intra,-}$: its phrase pairs $(x,y)$ have negative, low semantic similarity $\Phi(v^x, v^y)$.
\end{itemize}

We define a hybrid loss function as follows, which will be minimized to train the parameters (i.e., aggregation and matrix $\mathbf{W}^k$) of MGNN model:
\begin{equation}
    \mathcal{L} = \alpha\mathcal{L}_{1} + \beta\mathcal{L}_{2} + (1-\alpha-\beta)\mathcal{L}_{3}
\end{equation}
where $\alpha$ and $\beta$ are hyper-parameters and
\begin{equation}
\Scale[0.85]{\mathcal{L}_{1} = \sum_{\substack{(x_1, y_1) \in D_{meta,+} \\ (x_2,y_2) \in D_{intra,+}}} {\max \left( 0,\ \mathbf{z}^{x_2}\cdot \mathbf{z}^{y_2} - \mathbf{z}^{x_1}\cdot \mathbf{z}^{y_1} + \gamma_1 \right)},} \nonumber
\end{equation}
\begin{equation}
\Scale[0.85]{\mathcal{L}_{2} = \sum_{\substack{(x_1, y_1) \in D_{intra,+} \\ (x_2,y_2) \in D_{intra,-}}} {\max \left( 0,\ \mathbf{z}^{x_2}\cdot \mathbf{z}^{y_2} - \mathbf{z}^{x_1}\cdot \mathbf{z}^{y_1} + \gamma_2 \right)},} \nonumber
\end{equation}
\begin{equation}
\Scale[0.85]{\mathcal{L}_{3} = \sum_{\substack{(x_1, y_1) \in D_{intra,-} \\ (x_2,y_2) \in D_{meta,-}}} {\max \left( 0,\ \mathbf{z}^{x_2}\cdot \mathbf{z}^{y_2} - \mathbf{z}^{x_1}\cdot \mathbf{z}^{y_1} + \gamma_3 \right)},} \nonumber
\end{equation}
where $\gamma_i$ ($i=1,2,3$) are hyper-parameters.

\section{Experiments}
\subsection{Experimental Setup}

\subsubsection{Datasets}

ReVerb45K~\cite{vashishth2018cesi} has 45K relation tuples, 89K NPs, 21.6K RPs, and 106.4K sentences. The gold entities were obtained by linking NPs in the tuples to Freebase, resulting 7.5K gold entities. However, the dataset has no gold relation for canonical RPs. So we do quantitative analysis for NP canonicalization and qualitative analysis for RP canonicalization. We randomly sampled 20\% entities and used the associated tuples as the validation set. And the rest of the data was used for both unsupervised learning and test (i.e., test set).

\subsubsection{Evaluation Metrics}

Following~\cite{galarraga2014canonicalizing,vashishth2018cesi}, we use macro-, micro- and pairwise metrics for evaluating Open KB canonicalization methods. In all cases, F1 measure is given as the harmonic mean of precision and recall. 

\subsubsection{Baseline Methods}

For NP canonicalization, we compare MGNN with the following competitive methods: 

\noindent $\bullet$ \textbf{Morphological Normalization}~\cite{fader2011identifying} applies normalization operations; \textbf{Paraphrase Database (PPDB)} grouped two NPs together if they share a common paraphrase in PPDB 2.0~\cite{pavlick2015ppdb};

\noindent $\bullet$ Gal\'{a}rraga~\cite{galarraga2014canonicalizing} used IDF token similarity, Jaro-Winkler similarity metric and Attribute Overlap respectively, along with hierarchy clustering method to canonicalize OpenKB.

\noindent $\bullet$ \textbf{GloVe}~\cite{pennington2014glove} method represented NPs with pre-trained embeddings; 

\noindent $\bullet$ \textbf{HolE}~\cite{nickel2016holographic} has have been successfully applied for link prediction in KBs; 

\noindent $\bullet$ \textbf{CESI}~\cite{vashishth2018cesi} is a novel side information based embedding learning method for canonicalizing Open KBs. CESI solves a joint objective to learn noun and relation phrase embeddings, while utilizing relevant side information in a principled manner. CESI is now the state-of-the-art method on Open KB canonicalization. HolE is the main architecture of CESI, so we denote CESI as {HolE + Side Info}.


\subsection{Results on ReVerb45K}

\subsubsection{Overall Performance}
Table~\ref{tab:ReVerb_result} shows that the proposed MGNN outperforms all the competitive methods on the average result of the three evaluation metrics (i.e., macro-, micro- and pairwise-F1 score). Compared to GloVe, MGNN improves average F1 score relatively by 2.4\% (by 1.4\% on macro, by 0.6\% on micro, and by 5.2\% on pairwise, respectively). Investigating actual number of gold entities and precision/recall, MGNN successfully finds 225 more gold entities than GloVe and assigns 584 more NPs to the correct clusters. It significantly improves the precision of pairwise prediction (by 11.7\%). HolE only uses structure information to update embeddings, weakening the use of semantic information in Open KBs. Semantic embeddings (GloVe) are more effective to do NP canonicalization than HolE. MGNN aggregates both semantic and structural information, resulting the best performance.

The feature-based methods by Gal\'{a}rraga \textit{et al.}~\shortcite{galarraga2014canonicalizing} have competitive macro-F1 score but extremely low pairwise-F1. This is because most of the gold entities have very few NPs in the ReVerb45K, so they can be captured by the feature-based methods. These methods missed the gold entities that were frequently mentioned in the corpus like person names. Another reason is that ReVerb45K has a considerably large number of entities and a comparatively smaller number of relation tuples (89K vs 45K). These methods are more likely to put two NPs together if they share an uncommon token. So, the accuracy relies heavily on the quality of document frequency estimation though we may have a small number of tuples.

\begin{table}[t]
	\begin{center}
		\renewcommand{\arraystretch}{1.2}
		\setlength\tabcolsep{3pt}
		\begin{tabular}{|l|rrr|r|}
		\hline
		\bf Methods & \bf {Macro} & \bf {Micro} & \bf {Pair} & \bf {Aver.} \\ \hline \hline
		Gal\'{a}rraga-Attr & \underline{\textbf{75.1}} & 20.1 & 0.2 & 31.8\\
		Gal\'{a}rraga-StrSim & 69.9 & 51.7 & 0.5 & 40.7\\ 
        Gal\'{a}rraga-IDF & \underline{71.6} & 50.8 & 0.5 & 41.0\\
        Morph Norm &1.4 & 77.7 & 75.1 & 51.4 \\
        PPDB & 46.0 & 45.4 & 64.2 & 51.9\\ 
        HolE (Random) & 5.4 & 74.6 & 50.9 & 43.6\\
        HolE (GloVe) & 33.5 & 75.8 & 51.0 & 53.4\\ 
        GloVe & 56.3 & \underline{81.8} & \underline{77.0} & \underline{71.7}\\ 
        {MGNN (\textbf{Ours})} & 57.1 & \underline{\textbf{82.3}} & \underline{\textbf{81.0}} & \underline{\textbf{73.5}} \\
		\hline
		\hline
		{HolE (GloVe) + Side Info}& {62.7} & {84.4} & \underline{\textbf{81.9}} & {76.3} \\
		{MGNN + Side Info (\textbf{Ours})} & \underline{\textbf{66.7}} & \underline{\textbf{86.3}} & 81.2 & \underline{\textbf{78.3}} \\
		\hline
		\end{tabular}
	\end{center}
	\vspace{-0.1in}
	\caption{The MGNN model performs the best on NP canonicalization in ReVerb45k.}
    \label{tab:ReVerb_result}
    \vspace{-0.1in}
\end{table}

Side information could be useful as shown in previous work~\cite{vashishth2018cesi}, including WordNet, PPDB, and information obtained from entity linking and morph normalization. We implement a MGNN model equipped with the side information. It improves average F1 score relatively by 2.6\% over CESI (which is HolE being equipped with the side information). The MGNN model achieves a significantly bigger macro-F1 score (by +6.4\% over the best baseline). The MGNN achieves new state-of-the-art on ReVerb45k Open KB canonicalization.



\subsubsection{Ablation Study}
Table~\ref{tab:mg-gnn ablation} compares the variants of the proposed model to evaluate the effectiveness of the following components: (1) \textit{meta-graph based hybrid loss}, by discarding one of the three loss terms $L_i$ ($i=1,2,3$); (2) \textit{meta-graph based canonical embedding aggregation}, by removing the meta-graph based neighbor set $\mathcal{N}^{\Psi}$ in MGNN; (3) \textit{graph neural network architecture}, by discarding GNN and only using the proposed loss function to update semantic embeddings.

\begin{table}[t]
	\begin{center}
		\renewcommand{\arraystretch}{1.2}
		\setlength\tabcolsep{4pt}
		\Scale[0.95]{\begin{tabular}{|l|rrr|r|}
		\hline
		\bf & \bf {Macro} & \bf {Micro} & \bf {Pair} & \bf {Average} \\ \hline \hline
        {MGNN} & \underline{57.1} & \underline{\textbf{82.3}} & \underline{\textbf{81.0}} & \underline{\textbf{73.5}} \\ \hline \hline
        {w/o loss $\mathcal{L}_1$} & 56.1 & \underline{81.0} & \underline{79.5} & \underline{72.2}\\
        {w/o loss $\mathcal{L}_2$} & 41.3 & 74.6 & 68.4 & 61.4 \\
        {w/o loss $\mathcal{L}_3$} & 55.8 & 80.9 & 79.3 & 72.0\\
        {w/o loss $\mathcal{L}_1,\mathcal{L}_3$} & 54.9  & 80.3 & 77.6 & 70.9\\ \hline
        {w/o $\mathcal{N}^{\Psi}$} & 53.1 & 80.2 & 77.0 & 70.1\\ \hline
        {w/o GNN} & \underline{\textbf{58.3}} & 79.1 & 69.5 & 69.0 \\
		\hline
		\end{tabular}}
	\end{center}
	\vspace{-0.1in}
	\caption{Ablation study on the MGNN model.}
    \label{tab:mg-gnn ablation}
    \vspace{-0.1in}
\end{table}

\noindent \textbf{Meta-graph based hybrid loss:} We observe that MGNN with all the three loss terms consistently performs better than the models that discard any of them.
When $\mathcal{L}_2$ is missing, the performance is very poor and worse than the baselines. The reason is that the semantic similarity plays the most significant role in NP canonicalization though not perfect.
An additional loss term of $\mathcal{L}_1$ or $\mathcal{L}_3$ makes a relative F1 improvement by 1.5--1.8\%. 
Adding both $\mathcal{L}_1$ and $\mathcal{L}_3$ improves F1 relatively by 3.6\%.
It demonstrates the importance of the hybrid loss design, and indicates that the complementary of loss terms for learning canonical embeddings.

\noindent \textbf{Meta-graph based canonical embedding aggregation:} Given the first row (MGNN) and the 5th row (MGNN w/o $\mathcal{N}^{\Psi}$) in Table~\ref{tab:mg-gnn ablation}, one
can easily tell the improvement brought by the embedding aggregation from meta-graph based neighbors: relatively by 4.9\% on average F1.

\noindent \textbf{Graph neural network architecture:} Without using the GNN model, the average F1 score would drop relatively by 6.5\%. Moreover, the pairwise-F1 would drop relatively by 16.5\%. So the GNN model, which aggregates the context for each node into its canonical embedding, plays an important role in NP canonicalization. Without GNN, the macro-F1 is high but the micro- and pairwise-F1 are very low because ReVerb45K is sparse with a large number of entities and a comparatively small number of relation tuples.

\begin{table}[t]
	\begin{center}
		\renewcommand{\arraystretch}{1.1}
		\setlength\tabcolsep{4pt}
		\begin{tabular}{|l|}
		\hline
		\bf Two pure RP clusters \\ \hline
		\{\textit{announce acquisition of, acquire the asset of, }\\
		\textit{announce purchase of, become sole owner of, }\\
		\textit{buy control interest in, announce takeover of}\} \\ \hline 
		\{\textit{be the national language of, be the language} \\
		\textit{of, be the primary language use in, be speak in,} \\
		\textit{be the main language of, be an official ...}\} \\ \hline \hline
		\bf Two impure RP clusters \\ \hline
		\{\textit{be a citizen of, may have be bear in, }\\
		\textit{have grow up in, have be bear in}\\ \hline 
		\{\textit{be president of, be crown king of, become} \\
		\textit{leader of, become prime minister of}\} \\ \hline 
		\end{tabular}
	\end{center}
	\vspace{-0.1in}
	\caption{RP canonicalization cases on ReVerb45K.}
    \label{tab:relation}
    \vspace{-0.1in}
\end{table}

\subsubsection{Qualitative Analysis on RP Canonicalization}

We find a few interesting clusters in RP canonicalization. Table~\ref{tab:relation} presents examples. We have two types of clusters: (1) pure clusters where the RPs are well clustered and refer to one relation and (2) impure clusters where the RPs are not correctly clustered. In the pure clusters, we can see that MGNN can group RPs even though they have different surface forms, e.g., ``announce\_acquisition\_of'' and ``buy\_control\_interest\_in'', both referring to the \textit{acquisition} relation. Another example is ``be the language of'' and ``be speak in'', both referring to the \textit{native language} relation. In impure RP clusters, MGNN may be confused by relevancy: ``be\_a\_citizen\_of'' and ``have\_grow\_up\_in'' refer to relevant but not the same relation. Grouping ``be\_president\_of'' and ``become\_prime\_minister\_of'' is wrong though makes some sense. This needs to be addressed in future work.

\section{Related Work}
\noindent \textbf{Open IE:} OpenIE systems extend information extraction to open domains without requiring any relation-specific schema in advance~\cite{fader2011identifying,angeli2015leveraging,stanovsky2018supervised,jiang2019condition}. ReVerb~\cite{fader2011identifying} restricted the relation to verbal phrases. Early systems prefer to apply rule-based techniques to extract fact tuples~\cite{angeli2015leveraging}. Stanovsky \textit{et al.}~\shortcite{stanovsky2018supervised} obtained labeled OpenIE data from semantic role labeling.

\noindent \textbf{KB Canonicalization and Entity Linking:} Ontological KB canonicalization has been studied for long~\cite{krishnamurthy2011noun,pujara2013knowledge}. Concept Resolver took use of the ``one sense per category'' assumption which states that an entity mention refers to at most one concept in ontology~\cite{krishnamurthy2011noun}. Knowledge Graph Identification is to produce a consistent Knowledge Graph by performing entity resolution, entity classification, and link prediction jointly~\cite{pujara2013knowledge}. Pujara \textit{et al.}~\shortcite{pujara2013knowledge} incorporated multiple extraction sources and ontological information to infer the most probable knowledge graph. These approaches require additional information in the form of an ontology of relations, which is not available for Open KB. For Open KB canonicalization, Gal\'{a}rraga \textit{et al.}~\shortcite{galarraga2014canonicalizing} performed entity mention canonicalization over manually-defined feature spaces. Wu \textit{et al.}~\shortcite{wu2018towards} speeded up the canonicalization methods in practice.
Entity linking and named entity disambiguation aim at mapping entity mention to an existing KB such as Wikipedia or Freebase.
Most approaches~\cite{sil2018neural,raiman2018deeptype,murty2018hierarchical,ng2017machine} generated a list of candidate entities for each entity mention and re-rank them.


\noindent \textbf{Meta-Graph Analysis:} Zhao \textit{et al.}~\shortcite{zhao2017meta} first introduced the concept of meta-graph to heterogeneous information network to build recommender systems. They used meta-graph as features to measure the node similarity. Yang \textit{et al.}~\shortcite{yang2018meta} used meta-graph to learn the embedding of nodes in heterogeneous information networks. Most previous studies used meta-graph as a feature. We use meta-graph as an important structure indicating canonical properties in a multi-layered graph representation of Open KB. 

\section{Conclusions}
We proposed a multi-layered meta-graph based graph neural network model (MGNN) for Open KB canonicalization. MGNN integrates semantic information (intra-layer links) and structural information (inter-layer links) through canonical embedding aggregation. It adapted a meta-graph based neighbor acquisition and learned node canonical embedding with meta-graph based hybrid loss. Our model outperforms baselines on a general Open KB dataset.

\newpage
\balance
\bibliography{ref}

\begin{thebibliography}{}

\bibitem[\protect\citeauthoryear{Angeli \bgroup \em et al.\egroup
  }{2015}]{angeli2015leveraging}
Gabor Angeli, Melvin Jose~Johnson Premkumar, and Christopher~D Manning.
\newblock Leveraging linguistic structure for open domain information
  extraction.
\newblock In {\em ACL}, pages 344--354, 2015.

\bibitem[\protect\citeauthoryear{Devlin \bgroup \em et al.\egroup
  }{2019}]{devlin2018bert}
Jacob Devlin, Ming-Wei Chang, Kenton Lee, and Kristina Toutanova.
\newblock Bert: Pre-training of deep bidirectional transformers for language
  understanding.
\newblock In {\em NAACL}, 2019.

\bibitem[\protect\citeauthoryear{Fader \bgroup \em et al.\egroup
  }{2011}]{fader2011identifying}
Anthony Fader, Stephen Soderland, and Oren Etzioni.
\newblock Identifying relations for open information extraction.
\newblock In {\em Proceedings of the conference on empirical methods in natural
  language processing}, pages 1535--1545. Association for Computational
  Linguistics, 2011.

\bibitem[\protect\citeauthoryear{Gal{\'a}rraga \bgroup \em et al.\egroup
  }{2014}]{galarraga2014canonicalizing}
Luis Gal{\'a}rraga, Geremy Heitz, Kevin Murphy, and Fabian~M Suchanek.
\newblock Canonicalizing open knowledge bases.
\newblock In {\em Proceedings of the 23rd acm international conference on
  conference on information and knowledge management}, pages 1679--1688. ACM,
  2014.

\bibitem[\protect\citeauthoryear{Hamilton \bgroup \em et al.\egroup
  }{2017}]{hamilton2017inductive}
Will Hamilton, Zhitao Ying, and Jure Leskovec.
\newblock Inductive representation learning on large graphs.
\newblock In {\em Advances in Neural Information Processing Systems}, pages
  1024--1034, 2017.

\bibitem[\protect\citeauthoryear{Jiang \bgroup \em et al.\egroup
  }{2019}]{jiang2019condition}
Tianwen Jiang, Tong Zhao, Bing Qin, Ting Liu, Nitesh~V Chawla, and Meng Jiang.
\newblock The role of ``condition'': a novel scientific knowledge graph
  representation and construction model.
\newblock In {\em Proceedings of the 25th ACM SIGKDD International Conference
  on Knowledge Discovery \& Data Mining}. ACM, 2019.

\bibitem[\protect\citeauthoryear{Krishnamurthy and
  Mitchell}{2011}]{krishnamurthy2011noun}
Jayant Krishnamurthy and Tom~M Mitchell.
\newblock Which noun phrases denote which concepts?
\newblock In {\em Proceedings of the 49th Annual Meeting of the Association for
  Computational Linguistics: Human Language Technologies-Volume 1}, pages
  570--580. Association for Computational Linguistics, 2011.

\bibitem[\protect\citeauthoryear{Mikolov \bgroup \em et al.\egroup
  }{2013}]{mikolov2013distributed}
Tomas Mikolov, Ilya Sutskever, Kai Chen, Greg~S Corrado, and Jeff Dean.
\newblock Distributed representations of words and phrases and their
  compositionality.
\newblock In {\em Advances in neural information processing systems}, pages
  3111--3119, 2013.

\bibitem[\protect\citeauthoryear{Murty \bgroup \em et al.\egroup
  }{2018}]{murty2018hierarchical}
Shikhar Murty, Patrick Verga, Luke Vilnis, Irena Radovanovic, and Andrew
  McCallum.
\newblock Hierarchical losses and new resources for fine-grained entity typing
  and linking.
\newblock In {\em Proceedings of the 56th Annual Meeting of the Association for
  Computational Linguistics (Volume 1: Long Papers)}, pages 97--109, 2018.

\bibitem[\protect\citeauthoryear{Ng}{2017}]{ng2017machine}
Vincent Ng.
\newblock Machine learning for entity coreference resolution: a retrospective
  look at two decades of research.
\newblock In {\em Proceedings of the Thirty-First AAAI Conference on Artificial
  Intelligence}, pages 4877--4884. AAAI Press, 2017.

\bibitem[\protect\citeauthoryear{Nickel \bgroup \em et al.\egroup
  }{2016}]{nickel2016holographic}
Maximilian Nickel, Lorenzo Rosasco, and Tomaso Poggio.
\newblock Holographic embeddings of knowledge graphs.
\newblock In {\em Thirtieth Aaai conference on artificial intelligence}, 2016.

\bibitem[\protect\citeauthoryear{Pavlick \bgroup \em et al.\egroup
  }{2015}]{pavlick2015ppdb}
Ellie Pavlick, Pushpendre Rastogi, Juri Ganitkevitch, Benjamin Van~Durme, and
  Chris Callison-Burch.
\newblock Ppdb 2.0: Better paraphrase ranking, fine-grained entailment
  relations, word embeddings, and style classification.
\newblock In {\em Proceedings of the 53rd Annual Meeting of the Association for
  Computational Linguistics and the 7th International Joint Conference on
  Natural Language Processing (Volume 2: Short Papers)}, volume~2, pages
  425--430, 2015.

\bibitem[\protect\citeauthoryear{Pennington \bgroup \em et al.\egroup
  }{2014}]{pennington2014glove}
Jeffrey Pennington, Richard Socher, and Christopher Manning.
\newblock Glove: Global vectors for word representation.
\newblock In {\em Proceedings of the 2014 conference on empirical methods in
  natural language processing (EMNLP)}, pages 1532--1543, 2014.

\bibitem[\protect\citeauthoryear{Pujara \bgroup \em et al.\egroup
  }{2013}]{pujara2013knowledge}
Jay Pujara, Hui Miao, Lise Getoor, and William Cohen.
\newblock Knowledge graph identification.
\newblock In {\em International Semantic Web Conference}, pages 542--557.
  Springer, 2013.

\bibitem[\protect\citeauthoryear{Raiman and Raiman}{2018}]{raiman2018deeptype}
Jonathan~Raphael Raiman and Olivier~Michel Raiman.
\newblock Deeptype: multilingual entity linking by neural type system
  evolution.
\newblock In {\em Thirty-Second AAAI Conference on Artificial Intelligence},
  2018.

\bibitem[\protect\citeauthoryear{Saha}{2017}]{Saha2017BootstrappingFN}
IE~Swarnadeep Saha.
\newblock Bootstrapping for numerical open.
\newblock 2017.

\bibitem[\protect\citeauthoryear{Sil \bgroup \em et al.\egroup
  }{2018}]{sil2018neural}
Avirup Sil, Gourab Kundu, Radu Florian, and Wael Hamza.
\newblock Neural cross-lingual entity linking.
\newblock In {\em Thirty-Second AAAI Conference on Artificial Intelligence},
  2018.

\bibitem[\protect\citeauthoryear{Stanovsky \bgroup \em et al.\egroup
  }{2018}]{stanovsky2018supervised}
Gabriel Stanovsky, Julian Michael, Luke Zettlemoyer, and Ido Dagan.
\newblock Supervised open information extraction.
\newblock In {\em NAACL}, volume~1, pages 885--895, 2018.

\bibitem[\protect\citeauthoryear{Tan \bgroup \em et al.\egroup
  }{2006}]{tan2006introduction}
Pang-Ning Tan, Michael Steinbach, and Vipin Kumar.
\newblock Introduction to data mining.
\newblock 2006.

\bibitem[\protect\citeauthoryear{Vashishth \bgroup \em et al.\egroup
  }{2018}]{vashishth2018cesi}
Shikhar Vashishth, Prince Jain, and Partha Talukdar.
\newblock Cesi: Canonicalizing open knowledge bases using embeddings and side
  information.
\newblock In {\em Proceedings of the 2018 World Wide Web Conference on World
  Wide Web}, pages 1317--1327. International World Wide Web Conferences
  Steering Committee, 2018.

\bibitem[\protect\citeauthoryear{Wu \bgroup \em et al.\egroup
  }{2018}]{wu2018towards}
Tien-Hsuan Wu, Zhiyong Wu, Ben Kao, and Pengcheng Yin.
\newblock Towards practical open knowledge base canonicalization.
\newblock In {\em Proceedings of the 27th ACM International Conference on
  Information and Knowledge Management}, pages 883--892. ACM, 2018.

\bibitem[\protect\citeauthoryear{Yang \bgroup \em et al.\egroup
  }{2018}]{yang2018meta}
Carl Yang, Yichen Feng, Pan Li, Yu~Shi, and Jiawei Han.
\newblock Meta-graph based hin spectral embedding: Methods, analyses, and
  insights.
\newblock In {\em 2018 IEEE International Conference on Data Mining (ICDM)},
  pages 657--666. IEEE, 2018.

\bibitem[\protect\citeauthoryear{Zhao \bgroup \em et al.\egroup
  }{2017}]{zhao2017meta}
Huan Zhao, Quanming Yao, Jianda Li, Yangqiu Song, and Dik~Lun Lee.
\newblock Meta-graph based recommendation fusion over heterogeneous information
  networks.
\newblock In {\em Proceedings of the 23rd ACM SIGKDD International Conference
  on Knowledge Discovery and Data Mining}, pages 635--644. ACM, 2017.

\end{thebibliography}
\bibliographystyle{named}

\end{document}